\documentclass[sigconf, nonacm]{acmart}
\usepackage[ruled]{algorithm2e}
\usepackage{natbib}

\usepackage{booktabs} % For formal tables
\usepackage{subfigure}

\PassOptionsToPackage{table}{xcolor}

\AtBeginDocument{%
	\providecommand\BibTeX{{%
			\normalfont B\kern-0.5em{\scshape i\kern-0.25em b}\kern-0.8em\TeX}}}

\begin{document}
\title{Optimizing AD Pruning of Sponsored Search with Reinforcement Learning}
%% making the system more smooth
%Yijiang Lian, Zhijie Chen, Yifei Wang, Xin Pei, Shuang Li, Zhiheng Zhang, Zishuai Zhang, Zhipeng Tao, Yuefeng Qiu, Liang Yuan, Tianyu Wang, Ruiyu Yuan, Zhigang Li,

%\author{Yijiang Lian, Zhijie Chen, Xin Pei, Shuang Li, Yifei Wang, Yuefeng Qiu, Zhiheng Zhang, Zhipeng Tao, Liang Yuan, Hanju Guan, Kfeng Zhang, Zhigang Li, Xiaochun Liu}
%\affiliation{ \institution{Baidu} }
\author{Yijiang Lian$^1$, Zhijie Chen$^2$,Xin Pei$^1$, Shuang Li$^1$, Yifei Wang$^3$, Yuefeng Qiu$^1$, Zhiheng Zhang$^1$, Zhipeng Tao$^1$, Liang Yuan$^1$, Hanju Guan$^2$, Kefeng Zhang$^2$, Zhigang Li$^1$, Xiaochun Liu$^1$}
\affiliation{ \institution{$^1\ $\{lianyijiang, peixin, lishuang15, qiuyuefeng, zhangzhiheng02,taozhipeng, yuanliang, lizhigang01, liuxiaochun\}@baidu.com} }
\affiliation{ \institution{$^2\ $\{bytechen.hit, guanhanju, xidianzkf\}@gmail.com}}
\affiliation{ \institution{$^3$ doxa.wang@pku.edu.cn}}

%% \author{Yijiang Lian}
%% \affiliation{%
%%   \institution{Baidu}
%% }
%% \email{lianyijiang@baidu.com}

%% \author{Zhijie Chen}
%% \affiliation{%
%%   \institution{Baidu}
%% }
%% \email{bytechen.hit@gmail.com}

%% \author{Xin Pei}
%% \affiliation{%
%%   \institution{Baidu}
%% }
%% \email{peixin@baidu.com}

%% \author{Shuang Li}
%% \affiliation{%
%%   \institution{Baidu}
%% }
%% \email{lishuang15@baidu.com}

%% \author{Yifei Wang}
%% \affiliation{%
%%   \institution{Peking University}
%% }
%% \email{doxa.wang@pku.edu.cn}

%% \author{Zhiheng zhang}
%% \affiliation{%
%%   \institution{Baidu}
%% }
%% \email{zhangzhiheng02@baidu.com}

%% \author{Zishuai Zhang}
%% \affiliation{%
%%   \institution{Baidu}
%% }
%% \email{zhangzishuai@baidu.com}

%% \author{Zhipeng Tao}
%% \affiliation{%
%%   \institution{Baidu}
%% }
%% \email{taozhipeng@baidu.com}

%% \author{Liang Yuan}
%% \affiliation{%
%%   \institution{Baidu}
%% }
%% \email{yuanliang@baidu.com}

%% \author{Hanju Guan}
%% \affiliation{%
%%   \institution{Baidu}
%% }
%% \email{guanhanju@baidu.com}

%% \author{Kefeng Zhang}
%% \affiliation{%
%%   \institution{Baidu}
%% }
%% \email{xidianzkf@gmail.com}

%% \author{Zhigang Li}
%% \affiliation{%
%%   \institution{Baidu}
%% }
%% \email{lizhigang@baidu.com}

%% \author{Xiaochun liu}
%% \affiliation{%
%%   \institution{Baidu}
%% }
%% \email{liuxiaochun@baidu.com}

% The default list of authors is too long for headers.
%% \renewcommand{\shortauthors}{Yijiang Lian et al.}

\begin{abstract}
Industrial sponsored search system (SSS) can be logically divided into three modules:
keywords matching, ad retrieving, and  ranking.
During ad retrieving, the ad candidates grow exponentially. A query with high commercial value might retrieve
a great deal of ad candidates such that the ranking module could not afford.
Due to limited latency and computing resources, the candidates have to be pruned earlier.
Suppose we set a pruning line to cut  SSS into two parts:  upstream and  downstream.
The problem we are going to address is: how to pick out the best $K$ items from $N$ candidates provided by the upstream
to maximize the total system's revenue.
Since the industrial downstream is very complicated and  updated quickly,
a crucial restriction in this problem is that the selection scheme should get adapted to the downstream.
In this paper, we  propose a novel model-free reinforcement learning approach to fixing this problem.
Our approach considers  downstream as a black-box environment,
and the agent sequentially selects items and finally feeds into the downstream, where revenue
would be estimated and used as a reward to improve the selection policy. To the best of our knowledge, this is first time to
consider the system optimization from a downstream adaption view. It is also the first time to use
reinforcement learning techniques to tackle this problem. The idea has been successfully
realized in Baidu's sponsored search system, and online long time A/B test shows  remarkable improvements on revenue.
\end{abstract}

%%
%% The code below is generated by the tool at http://dl.acm.org/ccs.cfm.
%% Please copy and paste the code instead of the example below.
%%
%% \begin{CCSXML}
%% 	<ccs2012>
%% 	<concept>
%% 	<concept_id>10010520.10010553.10010562</concept_id>
%% 	<concept_desc>Information systems~Information retrieval</concept_desc>
%% 	<concept_significance>500</concept_significance>
%% 	</concept>
%% 	</ccs2012>
%% \end{CCSXML}

%% \ccsdesc[500]{Information systems~Information retrieval}

%% \keywords{Sponsored search, reinforcement learning, ad selection, global system optimization}

\maketitle
\section{Introduction}
\begin{figure}[h]
  \centering
    \includegraphics[width=0.4\textwidth]{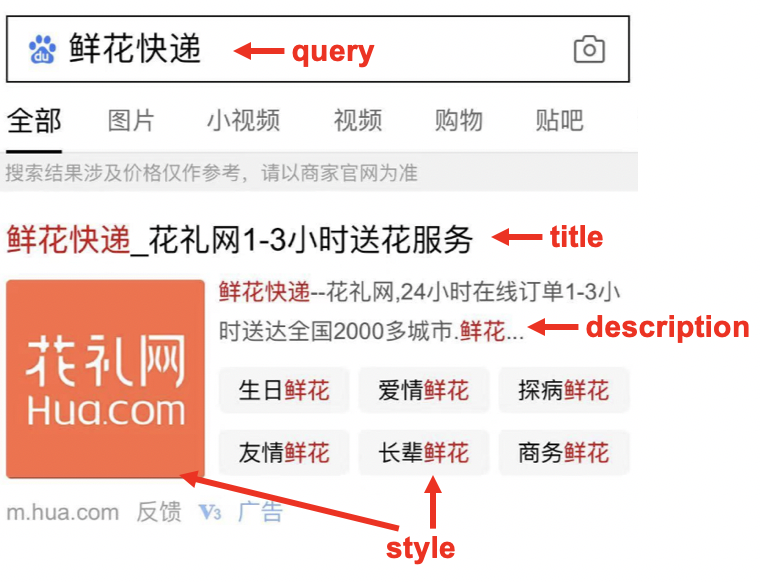}
  \caption{A typical shown ad has a title, a description and a style.}
  \label{fig:sponsored_ads}
\end{figure}

Web search engine plays a vital role in our daily life for seeking information. Since submitted
queries usually express a clear intent, search engine provides a good platform for advertisers to
accurately target clients. On this platform, advertisers need to submit keywords, bids and creatives
for their products and services. A \emph{Keyword} is a short text to be matched towards query traffic,
which can be seen as a bridge to link queries and ads. A \emph{Creative} is some texts that would be shown to users,
which generally includes a title and a description (Fig. \ref{fig:sponsored_ads}).
And the \emph{Bid} is used to express the advertiser's value for this keyword traffic.

%% elaborate matching mechanism(exact match, phrase match and broad match)
%% is provided by main sponsored search engines to satisfy different needs;

\begin{figure}[h]
  \centering
    \includegraphics[width=0.3\textwidth]{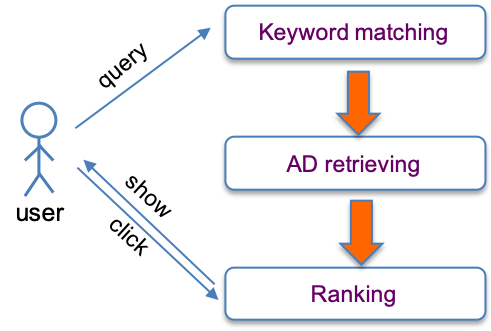}
  \caption{3 main modules in a sponsored search engine.}
  \label{fig:system_workflow}
\end{figure}

As shown in Fig. \ref{fig:system_workflow}, there are three main modules in a sponsored search engine (SSE):
keyword matching, ad retrieving and ranking. When a query is issued,
the \emph{keyword retrieving} module would retrieve all the semantically related keywords, and
the \emph{ad retrieving} module would retrieve all the ads of the advertisers who have bought the keywords, filter out
geographically or temporal conflicted ones and equip them with compatible styles.
Then, these fully equipped ads would be gathered to go through the \emph{ranking} module,
where lots of model predictions (like CTR (click through rate), relevance, etc.) and filtering rules are conducted,
then an auction (like GSP) is carried out for the remaining ads. Finally the winners would be shown to users.

A serious problem occurring in ad retrieving is that the candidates grow exponentially.
For example, a query with high commercial value
can retrieve 100 matching keywords, each keyword may be bought by 10 advertisers, then
each advertiser might design 10 different creatives for this unit, and 10 different display styles can be chosen. That is, we would get $10^5$ full ad candidates in this case. Directly sending these ads to the
\emph{ranking} module would consume a lot of computation time, which is unacceptable for an industrial online system. So we have to prune some candidates earlier during their expansion. Here we set our pruning line between the creative and the style equipping as shown
in Fig. \ref{fig:ad_assembling}. We refer to all the
modules below the pruning line as the downstream system.
Then a typical ad selection problem emerges: How to select $K$ items from $N$ candidates
to feed into the downstream system such that total revenue can be maximized?

There are several challenges for this problem.
Firstly, the real industrial
downstream system is very complicated, which usually
deals with ad CTR prediction, blacklist filtering, diversity filtering,
ad quality checking etc.
Without considering the downstream system, ads selected earlier might be filtered out a lot.
Secondly, at this moment, the ad candidate is not complete as style information is unavailable yet.
Thus we can not obtain a precise CTR estimation which is a key element for winning the auction.
Thirdly, this is a NP-hard combinatorial optimization problem.

Some heuristic approaches
are widely employed for tackling these challenges, such as sorting
all candidates under the revenue estimation and selecting the top $K$ items.
Obvious disadvantages of this kind of solutions are that, it is impossible to design perfect rules due to the super complexity of downstream system, and quality of the whole selection such as diversity is easily overlooked. Even if we adopt supervised learning based deep neural networks to solve it, it is hard to perform well as we are lack of best training samples.

In this paper, we resort to a reinforcement learning approach.
Using a model-free learning framework, the complex downstream  system can be considered as a black-box environment and the agent is an ad selector.
The agent sequentially selects $K$ items and feeds ad queue into the downstream system. The downstream system would estimate the final revenue for the ad queue, then this final eCPM (estimated cost per mille impressions) would be taken as a reward signal to guide the selection policy.
The merits of this approach are as  follows.
Firstly, the complex downstream system is fully taken into account so that we can optimize the system from a global view. Meanwhile, model-free learning frees us from understanding the complicated system, and the AD selection module can learn to smoothly get adapted
to the downstream system.
Secondly, we can use reinforcement learning's trial and error scheme to explore the better selections, and gradually improve our policy.

In summary, this paper offers the following main contributions:
\begin{enumerate}
\item We propose an ad pruning agent which adapts to the downstream system. We hope this idea would shed light on the future design of the industrial sponsored search system.
\item We propose a reinforcement learning based approach to tackling the ad selection problem, and this work has been successfully applied to Baidu's real sponsored search system.
\item For the concern of training efficiency and safety of the real online system, a simulator system for RL training has been devised and implemented.
\end{enumerate}

\section{Background}
\subsection{Ad retrieving}
\begin{figure}[h]
  \centering
    \includegraphics[width=0.35\textwidth]{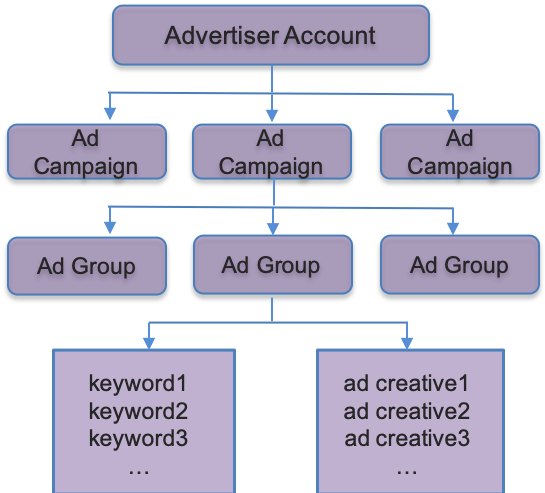}
  \caption{Sample of ads structure in advertiser accounts.}
  \label{fig:campaign_structure}
\end{figure}

Ads organization and retrieving process in Sponsor Search Systems (SSSs) are simply introduced in this section. An ad in SSS generally includes four components: \textit{keyword}, \textit{bid}, \textit{creative}, and \textit{style}.  Among them, the \textit{keyword}, \textit{bid}, and \textit{creative} are clearly defined by advertisers, while \textit{style} is either generated by our system or provided by advertisers. Ads in an advertiser account are hierarchically organized as shown in Fig. \ref {fig:campaign_structure}. From the top down, there are \textit{accounts}, \textit{campaigns}, \textit{groups} (or \textit{units}), and \textit{ads}. Flexible restrictions (e.g., target geographical area, time period, daily budget) can be easily set by advertisers at each level. Specially, ads in a single group are usually designed with a similar intention, so the keywords will be shared to match the retrieval traffic.

\begin{figure}[h]
  \centering
    \includegraphics[width=0.5\textwidth]{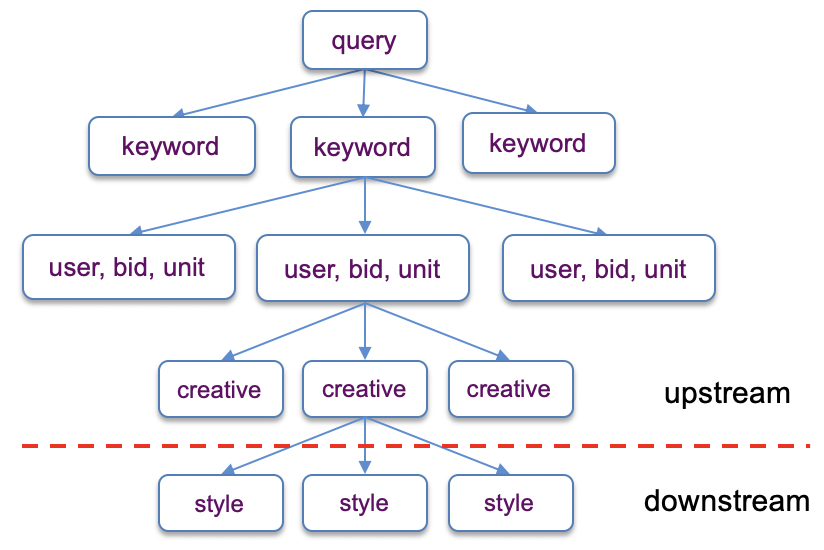}
  \caption{Candidates expand exponentially during ad retrieving.}
  \label{fig:ad_assembling}
\end{figure}

The ad retrieving process is carried out as shown in Fig. \ref{fig:ad_assembling}. When a query comes, matched keywords will be triggered out firstly through the Keyword Matching module. To make it feasible and efficient in practice, multi-phase retrieval is then adopted, and several Key-Value indexing sheets are constructed and saved. For example, given a keyword, a list of \textit{< (advertiser) user, unit>} pairs are firstly acquired with the inverted \textit{keyword-unit} indexing. If we expand the \textit{unit} with ads information, further \textit{<keyword, user, unit, creative>} will be obtained. Similarly, an ad creative can be expanded with various styles. At each phase we inquire the sheets, amount of candidates might be multiplied several times, and the full quantity would be significantly huge. More details about similar ad structure and retrieving process can be found in \cite{bendersky2010anatomy}. In this paper, our pruning agent acts after indexing ad creatives and before expanding styles.

\subsection{Terminology}
For clarity, we declare some terminology and notation in the following list:

\begin{enumerate}
\item SHOW denotes the total shown ad counts on result pages;
\item CTR denotes the average click ratio received by the search
  engine, which can be formalized as
  $\frac{\#\{\rm{clicks}\}}{\#\{\rm{searches}\}}$.
\item CPM denotes revenue received by search engine for 1000
  searches, which can be formalized as $\frac{\rm{revenue}}{\#\{\rm
    {searches}\}}\times 1000$.
\item eCTR denotes the estimated click through rate if an ad is shown.
\item eCPM denotes the estimated CPM. It equals to $\rm{eCTR}\times\rm{Charge}$.
\item Bid is the price provided by an advertiser for a keyword.
\item Charge is the true expense after the auction. Using GSP, charge is less than Bid.
\end{enumerate}

\section{Related Work}
% dl and rl for sponsored search
The combination of Deep Learning (DL) , known as Deep Reinforcement Learning (DRL), has led to great success, both in academic research and in industrial applications, such as games \cite{silver2016mastering}, finance \cite{heaton2017deep}, healthcare \cite{heaton2017deep}, as well as Google's machine translation system \cite{wu2016google}. Recently, the utilization of deep neural networks into sponsored search systems has yielded great benefits, particularly for matching queries and bidwords in the semantic space \cite{wu2018eenmf,gligorijevic2018deeply}. Advanced techniques such as generative sequence to sequence models have also been adopted to produce bidwords or match similar short sentences \cite{lee2018rare}. However, there are only a few existing works that incorporate RL/DRL techniques to sponsored search, e.g. \cite{zhao2018deep} for Real-Time Bidding. And as far as we know,  even few has been trialed for the ad pruning problem.

% ad pruning in Sponsored Search
In this work, we propose a Reinforcement Learning based approach to optimizing ad pruning under the adaptation consideration of the complicated and dynamic downstream system. As for the subset selection in ad pruning, it is a well-known NP-hard problem that requires exponential time to solve it exactly. Previous studies rely on various hand-designed heuristics to approximate the solution. Recent advances in deep learning provides an elegant and efficient method to such combinatorial optimization problems \cite{caldwell2018DeepOptimization,bengio2018CombinatorialOptimization}. In particular, we model the ad subset selection as a sequential decision making problem, and learns to maximize the overall reward of the selected subset, which is estimated by probing the downstream system. Similar optimization view to ours, \cite{buck2017ask} proposes an active question reformulation agent which interacts with a black box QA system and learns to reformulate questions to elicit best answers from the downstream.

\section{Methodology}

\subsection{Problem Formulation}

Suppose there are $N$ ad candidates provided by the upstream,
denoted as $X=\{x_1,x_2, \ldots, x_N\}$,
we are required to select best $K$ ads $Y=\{y_1, y_2, \ldots, y_K\}$
from $X$ and feed $Y$ to the downstream to maximize the revenue.

\begin{figure}[h]
  \centering
    \includegraphics[width=0.25\textwidth]{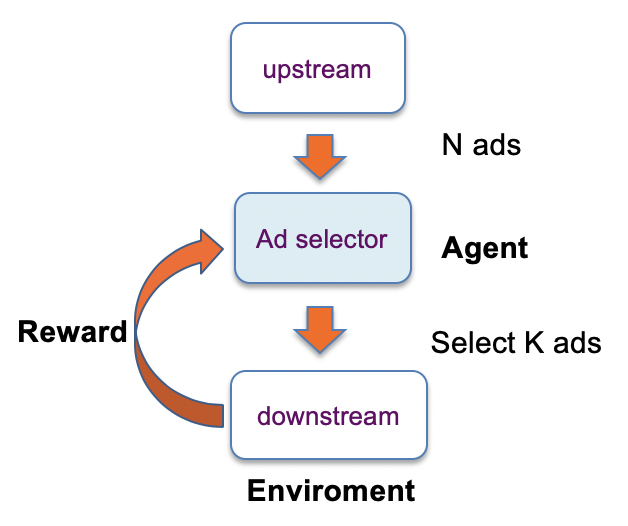}
  \caption{In our problem, the agent is the Ad selector, and the downstream is the
   environment.}
  \label{fig:agent_env}
\end{figure}

\begin{figure}[h]
  \centering
    \includegraphics[width=0.45\textwidth]{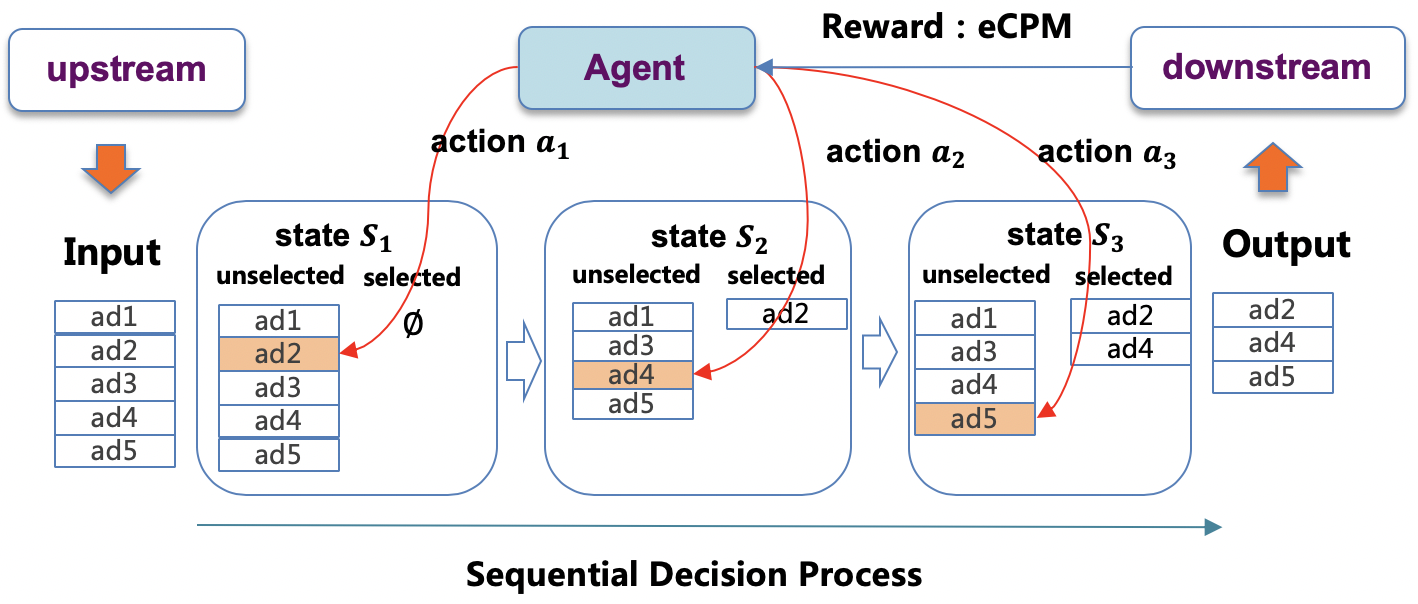}
  \caption{The sequential decision process of the pruning agent: The state comprises of the selected ads and unselected ones, and the action is to select one ad from the unselected set.
  In this illustration, the upstream provide $N=5$ ad candidates, and $K=3$ ads are picked out.
  At Step 1, the agent selects $ad2$, then $ad2$ is moved to the selected ad set. Similarly, $ad4$ and $ad5$ are selected at Step 2 and Step 3. Then
  output set $Y=\{ad2, ad4, ad5\}$ would be sent to the downstream.}

  \label{fig:sequential}
\end{figure}

We employ a model-free reinforcement learning approach for dealing with this problem. As shown in
Fig. \ref{fig:agent_env}, the complicated downstream system is treated as a black box environment,
and the agent sequentially selects $K$ ads with one ad at each step.
In particular, a Markov Decision Process is defined as follows (see Fig. \ref{fig:sequential}).
For each step $t$ ($1\leq t \leq K$) the state variable $s_t$ occupies
the current unselected ad list $X_t$ as well as the already selected ad list $Y_t$.
All ads in $X_t$ and $Y_t$ are drawn from the original set $X$. At start time, i.e. $t=1$, the unselected set equals to the origin input while the selected set is empty, i.e. $X_1=X$ and $Y_1=\varnothing$.
Based on the current state $s_t$,
each action $a_t$ is to choose one candidate ad from the unselected list $X_t$, and append it to the selected list.
The policy of our agent is a selection probability distribution over the whole unselected candidate set $X_t$. It is approximated by
a neural network parameterized by $\theta$, and denoted as $\pi_\theta(a_t|s_t)$.
When the selection is finished, these $K$ ads in $Y$ would be sent to the downstream system for further ranking and auction.
Finally, ads won out in this competition would be shown to users.
We denote the selection route as $\tau$, and the final winner ads as $Z=\{z_1, z_2, \ldots, z_M\}$. Then eCPM of $Z$
are used as the whole reward of this episode, that is,

\begin{equation}
\label{equ:equal_reward}
r(\tau)=\sum\limits_{i=1}^M\rm{eCPM}_{z_i}.
\end{equation}

The objective function we are going to optimize is:
\begin{equation}
\label{equ:rl_objective}
\max_\theta \; \big\{ L(\theta)=\mathbb{E}_{\tau\sim\pi_\theta}\left[r(\tau)\right] \big\}.
\end{equation}

\subsection{Training Algorithm}
We use Policy Gradient (PG) \cite{sutton2000policy} to solve the above problem. The optimization direction is decided by
\begin{equation}
\begin{split}
\label{equ:pg}
\nabla_\theta L(\theta)&=\nabla_\theta\mathbb{E}_{\tau\sim\pi_\theta}\left[r(\tau)\right] \\
&=\mathbb{E}_{\tau\sim\pi_\theta}\left[[\nabla_\theta\log \pi_\theta(a|s)]r(\tau)\right].
\end{split}
\end{equation}
and parameters are updated according to
\begin{equation}
\label{equ:para_update}
\theta \leftarrow \theta + \alpha\nabla_\theta L(\theta)
\end{equation}
where $\alpha$ is the learning rate.
The training procedure is explained in Algorithm \ref{alg:algo}.

%% For a single fake query, let it go throught the upstream, if the resulting ad queue's number
%% $>N$, then these candidates would go through the ad selector agent with parameter $\theta_n$
%% and pick out $K$ ads, the selector would log its selection path; then these $K$ ads would be send
%% to the downstream to get final reward $r(\tau)$, which would also be logged. Finally,
%% the selection path log $\tau$ and reward log $r(\tau)$ will join together to form a full training case
%% $(\tau, r(\tau))$. For a minibatch of queries, same process is conducted in parallel to get the training samples.
%% Then we use policy gradient to calculate the gradient $\nabla L$, and update the agent parameters to $\theta_{n+1}$.
\begin{algorithm} [h]
    \caption{Training algorithm}
    \newcommand{\Agent}{\textbf{Agent}}
        \While{{\rm{Agent's parameter $\theta$ is not converged}}}
        {
        sample a mini-batch of queries $Q$\;
        \For {{\rm{each query $q \in Q$} }}
        {
            \rm{go through the upstream to get $X_q$} \;
            \If {{ $|X_q| > K$}}
            {
                \textbf{Agent:} \rm{sequentially pick out $K$ ads $Y_q$ with parameter $\theta$} \;
                \rm{add selection path $\tau_q$ as $<q, \tau_q>$} to selection logs\;
                \rm{send $Y_q$ to the downstream} \;
                \textbf{Env:} calculate the reward $r(\tau_q)$  \;
                \rm{add reward $<q, r(\tau_q)>$} to reward logs\;
                \rm{join $<q, \tau_q, r(\tau_q)>$ } and add to training dataset\;
            }
        }
        \rm{sample (state, action, reward) from training dataset} \;
        \rm{update parameters $\theta$} according to Equ (\ref{equ:para_update}) \;
        }
     \label{alg:algo}
    \end{algorithm}
Specially,  we resort to our agent to pruning ads only if counts of the candidates provided by upstream is greater than $K$. The selection path $\tau_q$ by our agent would be logged in the form of $<q, \tau_q>$, and respective reward calculated by downstream is logged in the form of $<q, r(\tau_q)>$. The training samples are produced after joining these two logs.

%% \begin{equation}
%%     \theta^{(k+1)} = \theta^{(k)} + \alpha\frac{1}{M}\sum_{i=1}^M\sum_{t=1}^M\nabla_\theta\log\pi(b_t|s)r(B^{(i)}).
%% \end{equation}
%% Note the reward is only obtained after the sequential selection is completed, and hence the reward is sparse.

%% In practice we use on-policy samples to estimate the gradient. To reduce the variance of this estimation, we can remove a baseline $b(s)$ from the reward to reduce the variance
%% $$\nabla_\theta L=\mathbb{E}_{\tau\sim\pi_\theta}\left[[\nabla_\theta\log \pi_\theta(a|s)](r(s,a)-b(s))\right]$$

%\subsection{Reward shaping}

One more problem is that the agent takes $K$ selecting actions for the downstream, but only one reward for the whole sequence is available. If we make no distinction between these actions and give equal reward for each selection, the training will be inefficient. Here we adopt reward shaping \cite{ng1999policy} is to overcome this shortcoming, which has been applied in
Machine Translation \cite{bahdanau2016actor,wu2018study}. In our scenario,
the reward is set according to the eCPM contributions.
It means that the reward for selecting a shown ad $z \in Z $ is set to $\rm{eCPM}_z$,  while the reward for selecting an unshown ad is zero.
Additionally, since the eCPM differs a lot among different ads, to reduce this variance, logarithm transformation is adopted to smooth the original eCPM.

\subsection{Training Architecture}

People may argue that: why not use the real CPM as the reward rather than eCPM. RL algorithms commonly require a large number of interactions with the environment. On one hand using real CPM means that we have to wait for the real user's feedback, which may take a long while. And on the one other hand policy exploration on the real traffic probably greatly damnify the daily revenue especially for a commercial system.
For efficiency and safety concern, we have built an industrial sponsored search system simulator for our training, which can offer us a reliable and dense reward estimation without doing any harm to actual revenue.

\begin{figure}[h]
	\centering
	\includegraphics[width=0.45\textwidth]{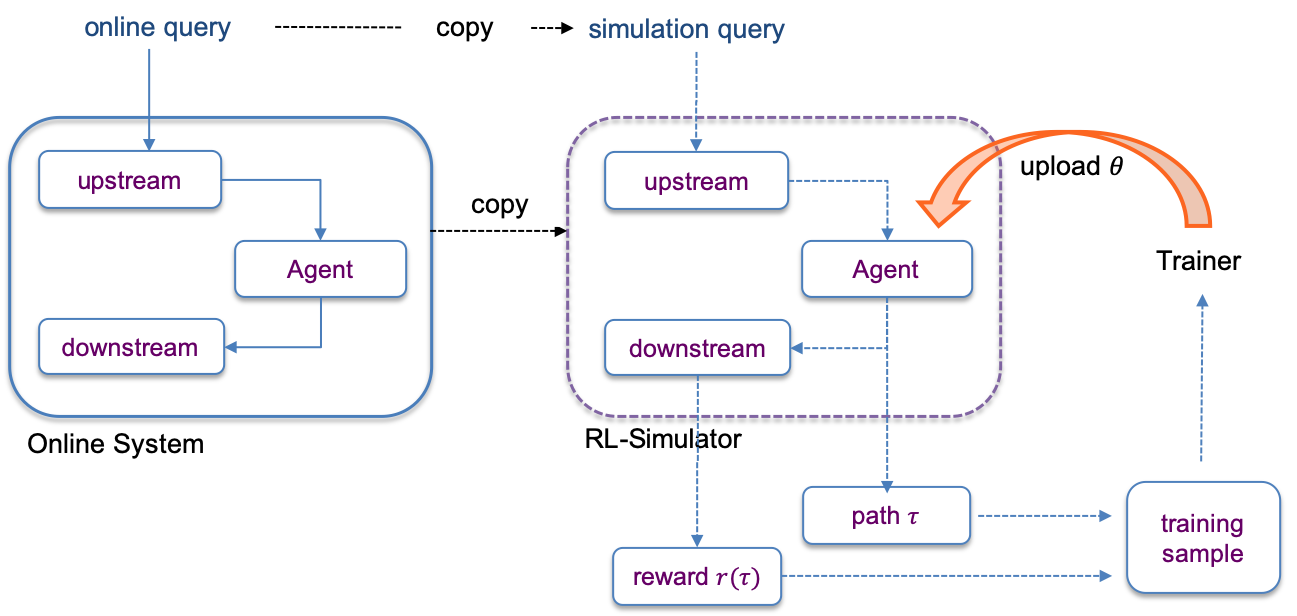}
	\caption{Training architecture with simulator.}
	\label{fig:simulator_architecture}
\end{figure}

Fig. \ref{fig:simulator_architecture} illustrates the training architecture with the simulator. Firstly, the whole system has been cloned as a simulation environment, and each query issued to the real online system will be copied to the simulated upstream. Secondly, the simulator agent will exploringly select ads according to the current policy, action trajectory of the selection procedure as well as the  estimated reward by the simulated downstream will be logged and stored as training samples. Thirdly, the offline trainer update the policy parameters with the training data, and push to the simulator agent in nearly real-time. With such a training architecture, the online system, simulator system, and the trainer are decoupled, high-throughput and low-latency for efficient training can be assured.

Before we publish a trained model to the online agent, to minimize the possible negative impact on the online system, two steps of model evaluations are also taken. In the first step, we check the performance of the model policies on the simulator environment by eCPM. And in the second, CPM is evaluated on the online system with a small fraction of real traffic. Only models perform well at both steps can be used in the entire online system.

\section{Experiments}
\subsection{Setup}

Experiments are conducted on Baidu's sponsor search system. The model to approximate the candidate ad selection score at each step is a two-layer fully connected network. The input layer consists of designed features, and the output layer generates one-dimensional real values as ad scores. Final selection probabilities are obtained through a \textit{softmax} layer. During the training,  actions are taken according to the probability distribution for policy exploration, while during  the online inference, only the candidate with maximum probability is chosen for the best exploitation.

More specifically, in our experiment, the maximum number $N$ of candidates  is set to 1000, and the selection count $K$ is at most $100$.
Through grid search of related parameters, batch size during training is set to 128, and for each query $q$ in the batch,
we sample 50 states from the decision route $\tau_q$. Adam optimizer is adopted
to update the agent parameters with $\alpha=0.99$, $ \beta=0.999$, and
initial learning rate $10^{-4}$.

For features, two categories for each candidate ad are designed: static and dynamic. Static features describe either the ad itself, or the query/user itself, or ad-query properties, and dynamic features characterize attributes related to the already selected ad collection. That is, the difference lies in whether it changes with the sequential selecting process.  The dynamic features aim to help maximize cumulative revenues, as well as maintain the diversity of advertisers. Main features we used are listed in Table \ref{tab:feature}.

\begin{table}[h]
\centering
\begin{tabular}{ll}
\toprule
\textbf{Type} & \textbf{Feature Name} \\
\toprule
static & estimated click through rate  \\
static & query-ad relevance \\
static & bid \\
static & pre-trained query embedding \\
static & pre-trained ad embedding \\
static & if this ad is compatible with strong style \\
dynamic & if the same advertiser's ad has been selected \\
dynamic & accumulated eCPM within same advertiser \\
\bottomrule
\end{tabular}
\caption{The main features.}
\label{tab:feature}
\end{table}

\subsection{Baselines}

We compare our approach with two baselines. They are commonly used in industry and also adopted in our sponsored search system before this method.
The first ranking baseline is by expected charge from the advertiser which equals to estimated click-through rate (eCTR) times bid. The second is ranked by estimated achievement from the downstream system, which is calculated by expected charge times the show probability (srq). The show probability is added due to the limited advertising position and multiple factors effect the eventual show results. Here we approximate the show probability of an ad to its accumulated show proportion in the past 30 days. In the following section, we denote these two baselines by \textbf{eCTR*bid } and \textbf{ eCTR*bid*srq}.

\subsection{Results}

We show the key online A/B testing results in Table \ref{tab:ab_res}. Three most important indicators of our method are compared with the baselines. They are click count per thousand searches (CTR) , shown ad count per thousand searches (SHOW), and revenue per thousand searches (CPM). The results demonstrate that our method achieves positive improvements over all three concerned indicators with both the baseline methods. Considering that ad supplies are stable, increment in SHOW shows that our agent does select better ad candidates which are more compatible with the downstream system. In addition, the proposed method also gains significant CTR improvements, namely, 1.11\% over \textbf{eCTR*bid*srq } and 2.21\% over \textbf{eCTR*bid}. It denotes that the newly shown ads are greatly accepted by users. Owing to the consistency of our pruning agent and the down-stream system, as well as the end user preference, we finally achieve a dramatic 1.95\% improvement of CPM.

\begin{table}[h]
\centering
\begin{tabular}{llll}
\toprule
\textbf{baselines} & \textbf{CTR} & \textbf{SHOW} & \textbf{CPM} \\
\toprule
\textbf{eCTR*bid} & 2.21\%  & 1.27\% & 1.95\% \\
\textbf{eCTR*bid*srq} & 1.11\% & 1.17\% & 0.99\% \\
\bottomrule
\end{tabular}
\caption{Online A/B Test Results.}
\label{tab:ab_res}
\end{table}

\begin{figure}[htbp]
	\centering
	\subfigure[Training loss with training steps (k)]{
		\begin{minipage}[t]{0.5\linewidth}
			\centering
			\includegraphics[width=1.75in]{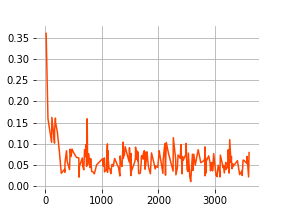}
			%\caption{fig1}
		\end{minipage}%
	}%
	\subfigure[Recall of top ctr*bid ads with training steps (k)]{
		\begin{minipage}[t]{0.5\linewidth}
			\centering
			\includegraphics[width=1.75in]{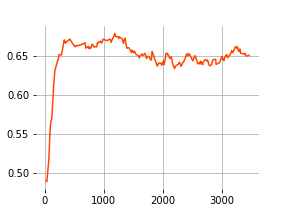}
			%\caption{fig2}
		\end{minipage}%
	}%
	\centering
	\caption{Training curves}
	\label{fig:training_curves}
\end{figure}

To illustrate the training performance of our pruning agent, we observe the training curves of our agent as well as its online performance with different training steps. Fig. \ref{fig:training_curves}(a) displays the training loss, and Fig. \ref{fig:training_curves}(b) plots the recall of top expected charge candidates (top 110 eCTR*bid). Both the decrease in training loss and increase in recall meet our expectation, and convergence is reached around 1500k steps.
More specifically, the best recall of top eCTR*bid candidates is around 0.67 rather than 1, which demonstrates the imparity between our method and the \textbf{eCTR*bid} baseline.
Furthermore, we choose several optimization versions after various training steps. Table \ref{tab:ab_res_with_steps} shows the online performance in comparison with the \textbf{eCTR*bid} baseline. We can observe that the indicators improves with training steps, and the best online performance is achieved after 2000k steps, which validates the effectiveness of our model training.

\begin{table}[h]
	\centering
	\begin{tabular}{llll}
		\toprule
		\textbf{Training Steps} & \textbf{CTR} & \textbf{SHOW} & \textbf{CPM} \\
		\toprule
		\textbf{1} & -2.90\%& -1.68\% & -11.20\% \\
		\textbf{500k} & -0.13\% & 0.47\% & 0.54\% \\
		\textbf{1000k} & 1.81 \% & 1.03 \% & 1.30\% \\
		\textbf{1500k} & 2.02\% & 1.21\% & 1.61\% \\
		\textbf{2000k} & 2.23\%  & 1.26\% & 1.93\% \\
		\textbf{2500k} & 2.21\%  & 1.27\% & 1.95\% \\
		\bottomrule
	\end{tabular}
	\caption{Online performance with training steps}
	\label{tab:ab_res_with_steps}
\end{table}

It is also worth mentioning that this agent assumes the downstream system to be a completely black box, whereas it may be not. For example, in our system, it is clear that the downstream part would select at most $x$ ads from a same advertiser. Even though that our model succeeds in learning to select diverse ads based on the given features and rewards, we attempt to directly apply this prior knowledge during both the training and online inferring. The result shows that this application brings us a significant return of performance (that is nearly saving 10\% time in selecting actions) while without lessening the CPM revenue. We believe that making full use of similar prior knowledge in modeling is a valuable practice in industry systems.

\section{Concluding Remarks}

In this paper we have devised a reinforcement learning method to address the ad pruning problem
in the real sponsored search engine environment. During ad retrieving, the ad candidates expands
exponentially. To meet the latency requirements, these candidates need to be pruned earlier.
An ad selection agent is set at the pruning point, and cuts the system into upstream and downstream.
Without concerning the downstream, the selected ads might be filtered later. To address this problem,
we have considered the complicated downstream as a black box environment, and our agent sequentially selects
$K$ ads and feeds them into the downstream to get the eCPM reward for training. Online long term A/B test on Baidu's sponsored search engine has showed that it greatly outperforms the CPM rule based selection approaches.
A similar reinforcement learning based method has been applied in other Baidu products, which also yields great improvements.

%\section{Future work}

In our mind, making ad selection adapted to the downstream system is crucial. This
\emph{downstream adaptation} method can be applied in many other online services. For example, most online search or
recommendation service has a coarse-ranking and refined-ranking part, and the coarse-ranking's job is to select limited candidates from a vast mount of items and feed them into the downstream part. Our method can be easily transplanted to these scenarios.

%
% The next two lines define the bibliography style to be used, and the bibliography file.
\section{acknowledgment}
We would like to thank Tianyu Wang, Zishuai Zhang, Ruiyu Yuan's efforts in applying this RL method to different ad product scenarios. And we would also like to thank Yanjun Jiang and Tefeng Chen for their construction of the environment simulator.
\bibliographystyle{ACM-Reference-Format}
\bibliography{sample-base}
\end{document}